\documentclass[11pt,twocolumn]{article}
\usepackage{times}
\usepackage[pdftex]{graphicx}
\usepackage{amssymb, amsmath}
\usepackage{fullpage}

\begin{document}

\title{\vspace{1cm}PLU: The Piecewise Linear Unit Activation Function}
\author{Andrei Nicolae\footnote{nandrei@u.washington.edu} \\ University of Washington, Seattle}
\date{\vspace{1cm}}

\maketitle

\begin{abstract}
Successive linear transforms followed by nonlinear ``activation" functions can approximate nonlinear functions to arbitrary precision given sufficient layers. The number of necessary layers is dependent on, in part, by the nature of the activation function. The hyperbolic tangent (tanh) has been a favorable choice as an activation until the networks grew deeper and the vanishing gradients posed a hindrance during training. For this reason the Rectified Linear Unit (ReLU) defined by $max(0, x)$ has become the prevailing activation function in deep neural networks. Unlike the tanh function which is smooth, the ReLU yields networks that are piecewise linear functions with a limited number of facets. This paper presents a new activation function, the Piecewise Linear Unit (PLU) that is a hybrid of tanh and ReLU and shown to outperform the ReLU on a variety of tasks while avoiding the vanishing gradients issue of the tanh. 
\end{abstract}

\section{Introduction}
When a linear function $h(x)$ is transformed by the hyperbolic tangent, i.e. $g(x) = \tanh(h(x))$, the resulting function $g(x)$ is nonlinear and smooth. When the ReLU is likewise applied to $h(x)$, the result is a piecewise linear function with derivative either $0$ or $\nabla h$. Approximating a smooth, highly nonlinear function using a network model requires many ReLU activations, implying a network with many layers. To increase the nonlinearity created by each activation, the Piecewise Linear Unit (PLU) activation function is proposed:

\begin{equation*}
PLU(x) \equiv max(\alpha(x+c)-c, min(\alpha(x-c)+c, x))
\end{equation*}

\begin{figure}[t]
\centering
\includegraphics[width=.5\textwidth]{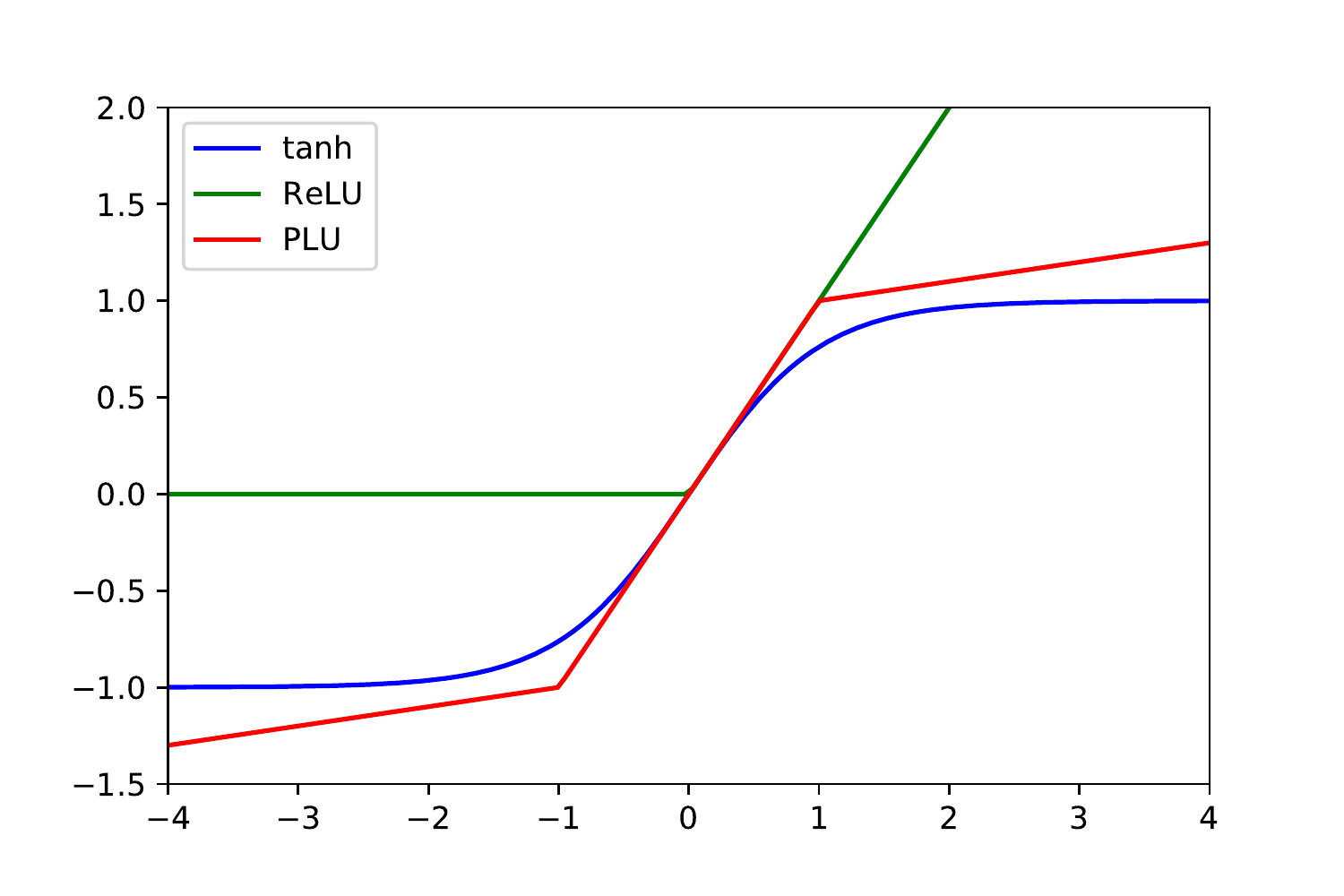}
\caption{Plot of the tanh, ReLU, and PLU activation functions. For the PLU, $\alpha=0.1, c=1$.}
\label{activations}
\end{figure}

The PLU is a crude piecewise approximation of the tanh and is unbounded in order to avoid vanishing gradients. The ReLU function is made of two linear segments, with gradients 0 or 1. The PLU has three linear segments with gradients $\alpha$ or $1$, where $\alpha$ is a parameter to be chosen or trained. As it will be shown, the PLU can fit highly nonlinear functions more closely than the ReLU. 

\section{Fitting Elementary Functions} \label{sec:sine}
A network model $F: \mathbb{R} \to \mathbb{R}$  was constructed to test the three activation functions in figure (\ref{activations}).
\begin{equation*}
F(x) = W_3 \ a(W_2 \ a(W_1x + b_1) + b_2) + b_3
\end{equation*}
This defines a network of depth 3 and width 3, i.e. $W_2 \in \mathbb{R}^{3x3}$. The function $a: \mathbb{R} \to \mathbb{R}$ is one of the three tested activations, applied element-wise. 

This network was implemented in TensorFlow and trained using the Adam optimizer with a learning rate of 0.01 and default parameters otherwise. The function to fit was $f(x) = \sin(x)$, $x \in [-2\pi, 2\pi]$. The $(x,y)$ training data was fed as 50 linearly spaced points on the domain of $x$, i.e. \texttt{x = np.linspace(-2 * np.pi, 2 * np.pi, 50)} along with their true function values $y = \sin(x)$. The parameters $W_i$ were initialized randomly from $\mathcal{N}(0,1)$, while $b_i$ initialized to zeros. The network was trained for 2048\ steps.

\begin{figure}[t]
\centering
\includegraphics[width=.5\textwidth]{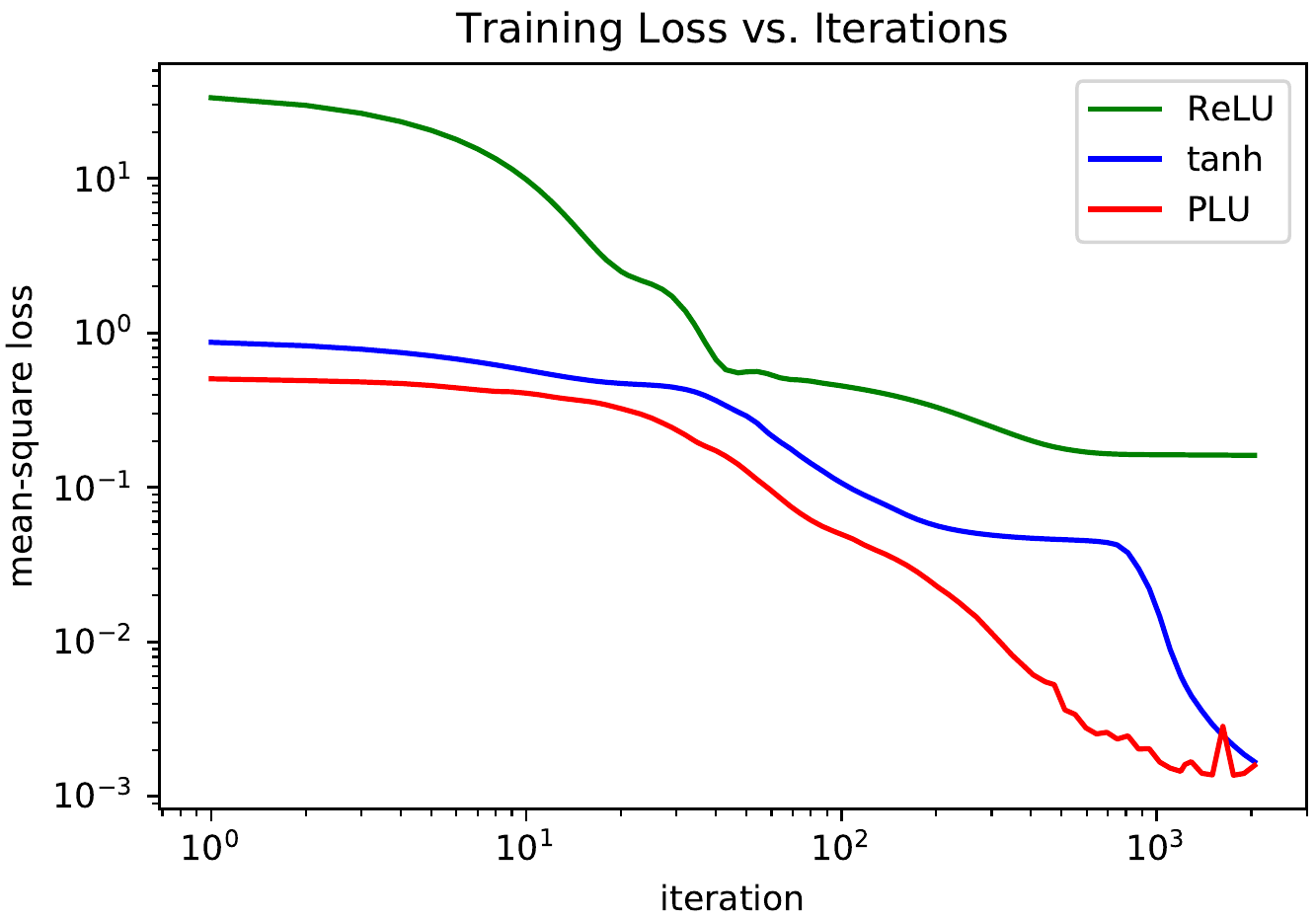}
\caption{Loss function for all three activations.}
\label{sineLoss}
\end{figure}

\begin{figure}[h!]
\centering
\includegraphics[width=.5\textwidth]{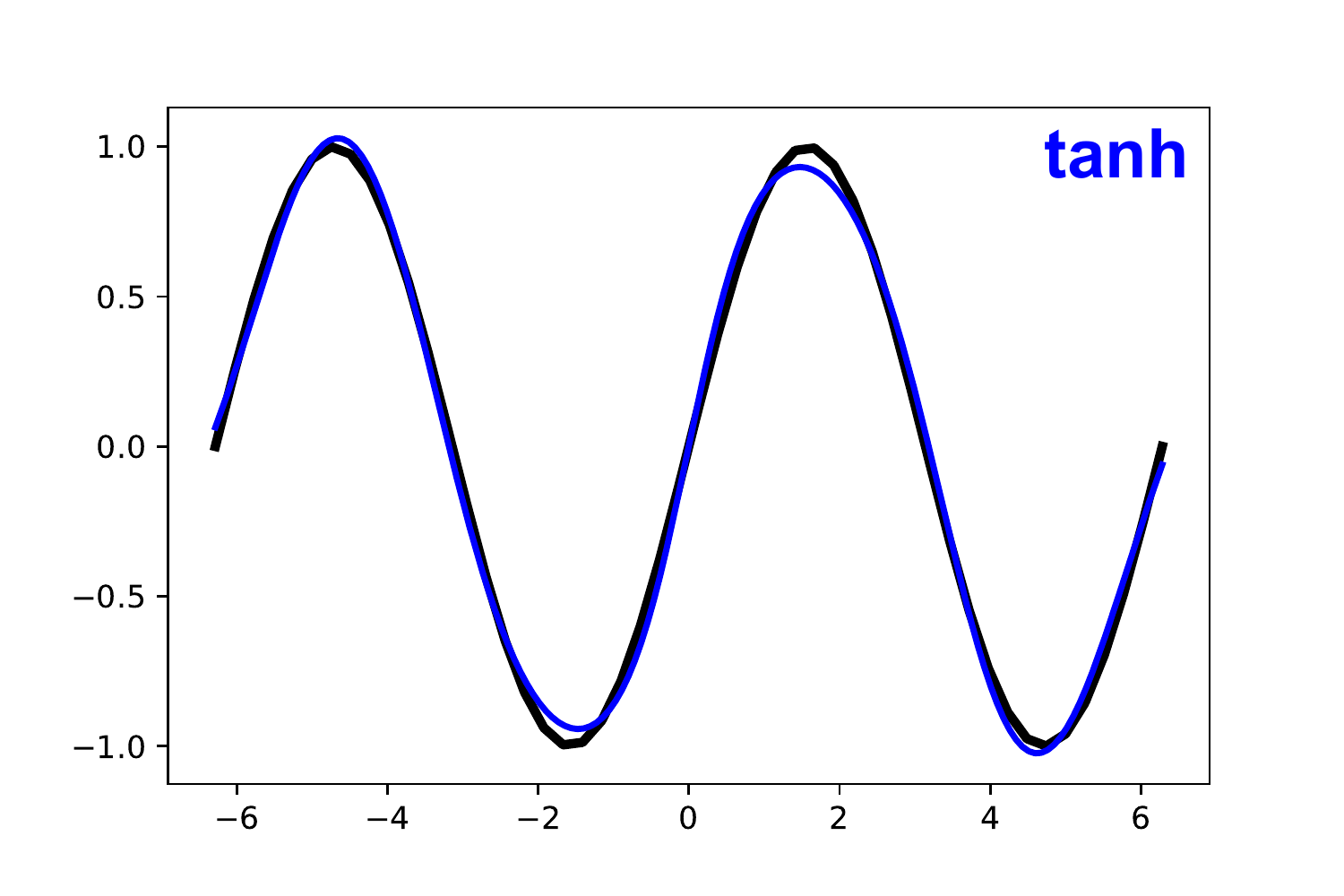}
\includegraphics[width=.5\textwidth]{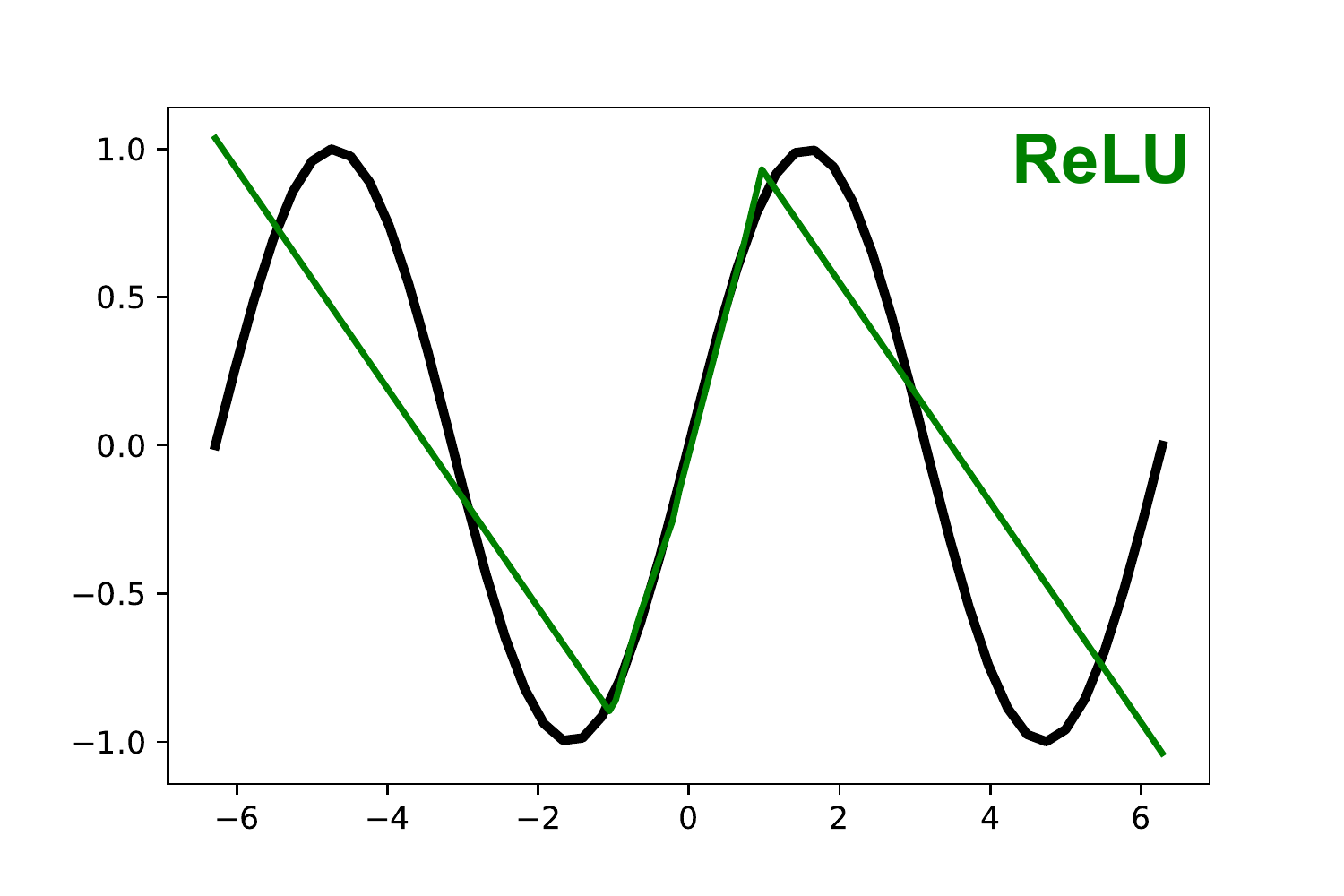}
\includegraphics[width=.5\textwidth]{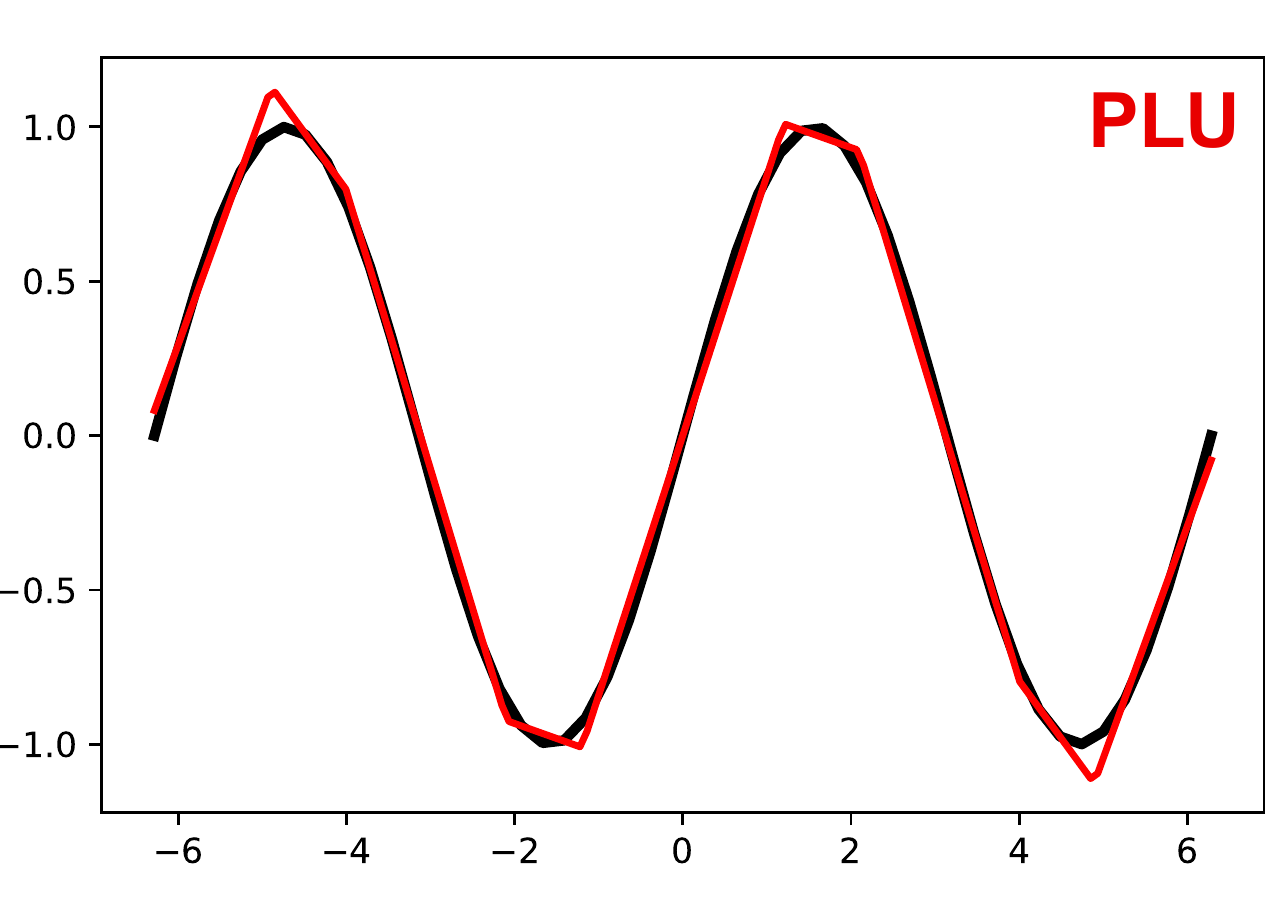}
\caption{$\sin(x)$ function approximated by a neural network using three different activation
\label{sinePlots} functions.}
\end{figure}

Function evaluations of $F(x)$ (the network) and $\sin(x)$ (black line) are shown in figure (\ref{sinePlots}). Note the smoothness of the network model activated by tanh, and the piecewise linear nature of both ReLU and PLU. There is a difference in the number of line segments composing $F(x)$ between the ReLU and PLU due to their definitions. Since the network is shallow and the activation only applied twice, the ReLU network could not fit the contour of the sine function since it had an insufficient number of pieces. Thus, the loss had a higher lower bound than the PLU and tanh, as shown in the figures. Figure (\ref{sineLoss}) shows the mean square loss of all three networks during training. The PLU and tanh converge to a similar MSE loss, two orders of magnitude lower than the ReLU.

\section{Fitting Parametric Functions}
Next, the network's depth and width were both increased to 5 and the output has dimension 2, i.e. $F: \mathbb{R} \to \mathbb{R}^2$. The new function to fit is parametric, defined by:
\begin{equation*}
f(t) = [(\cos(c_1t) -  \cos(c_2t))^3, (\sin(c_3t) -  \sin(c_4t))^3]
\end{equation*}
with $\vec{c} = [1, 2, 2, 1]$. This network was trained for 4096 steps for each activation - all other hyperparameters were the same as before.

\begin{figure}[h!]
\centering
\includegraphics[width=.412\textwidth]{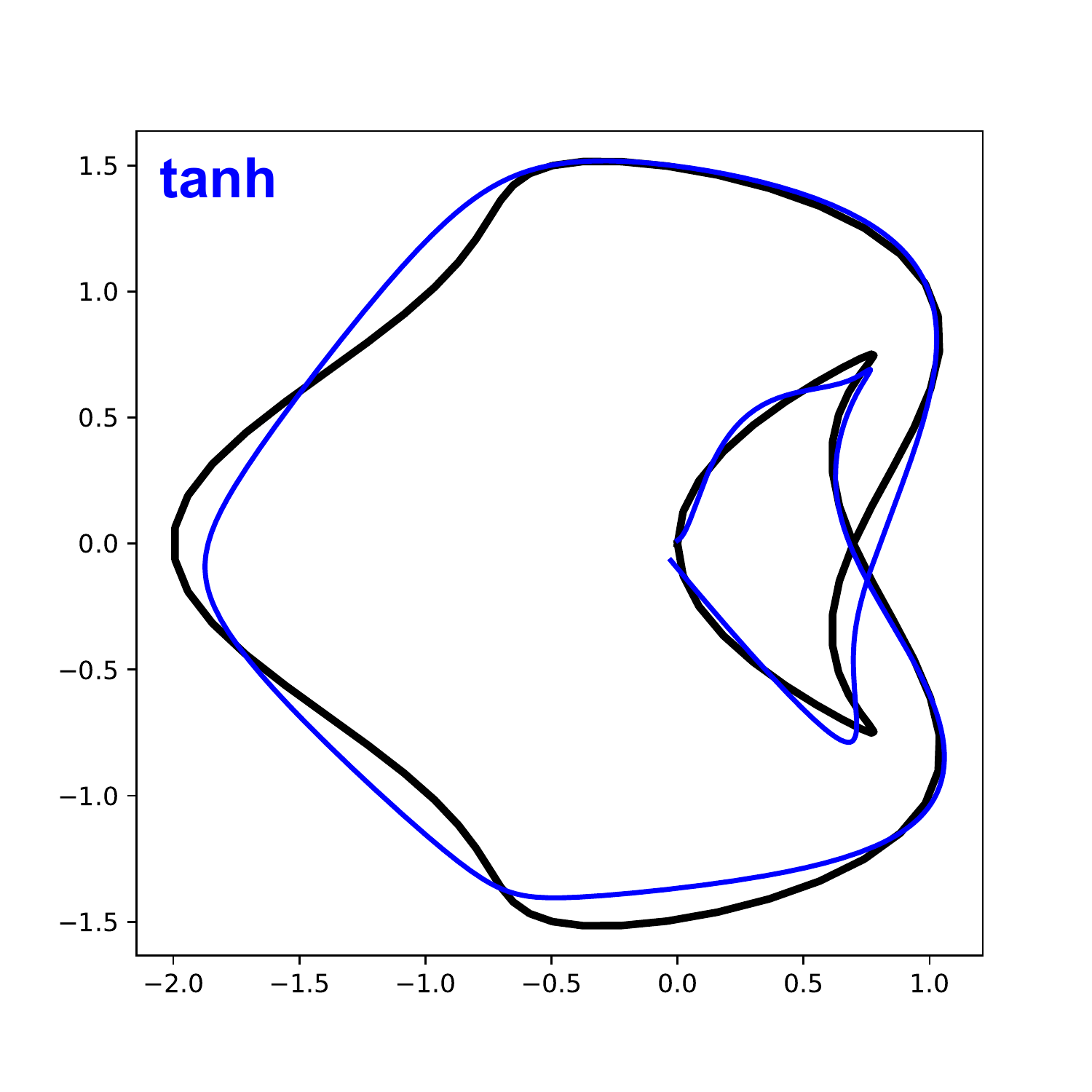}
\includegraphics[width=.412\textwidth]{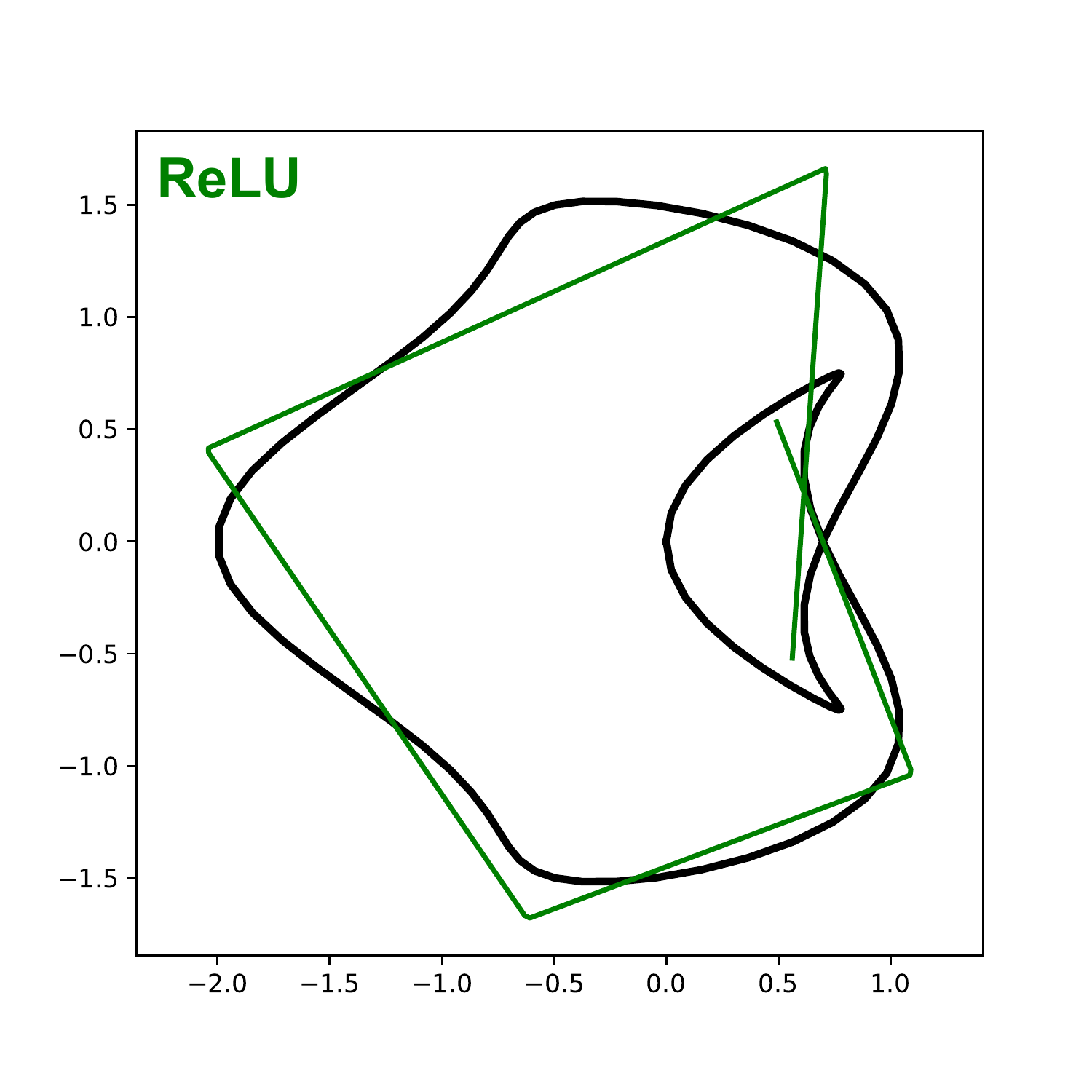}
\includegraphics[width=.412\textwidth]{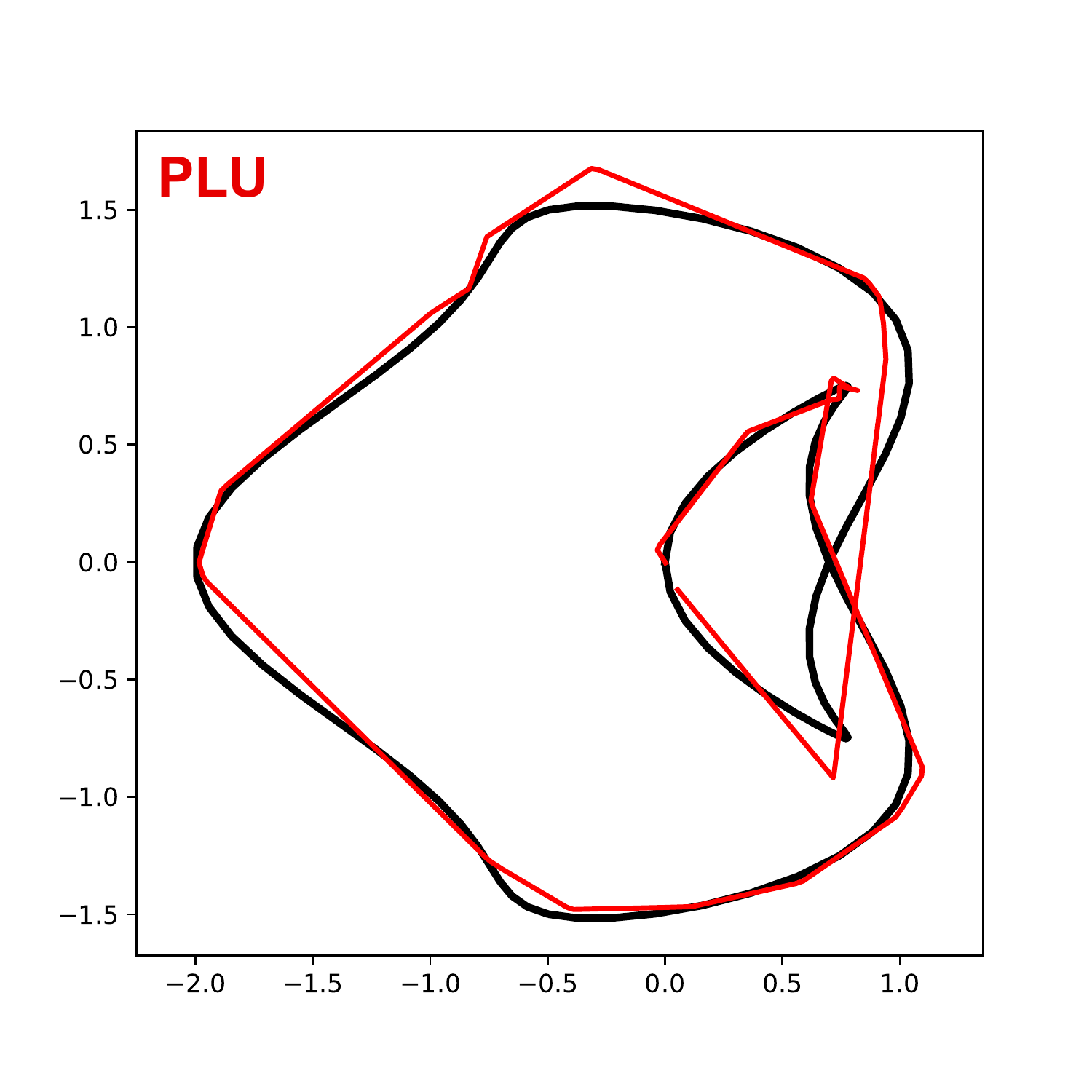}
\caption{Parametric function approximation.}
\label{parametricPlots}
\end{figure}

Plots of $F(t)$ (color) and $f(t)$ (black) are shown in figure (\ref{parametricPlots}). Again, the ReLU struggles to approximate $f(t)$ with its 5 line segments while the PLU has abundantly more, thus fitting much closer. The tanh also performs well and is smooth as expected. 
\begin{figure}[t]
\centering
\includegraphics[width=.5\textwidth]{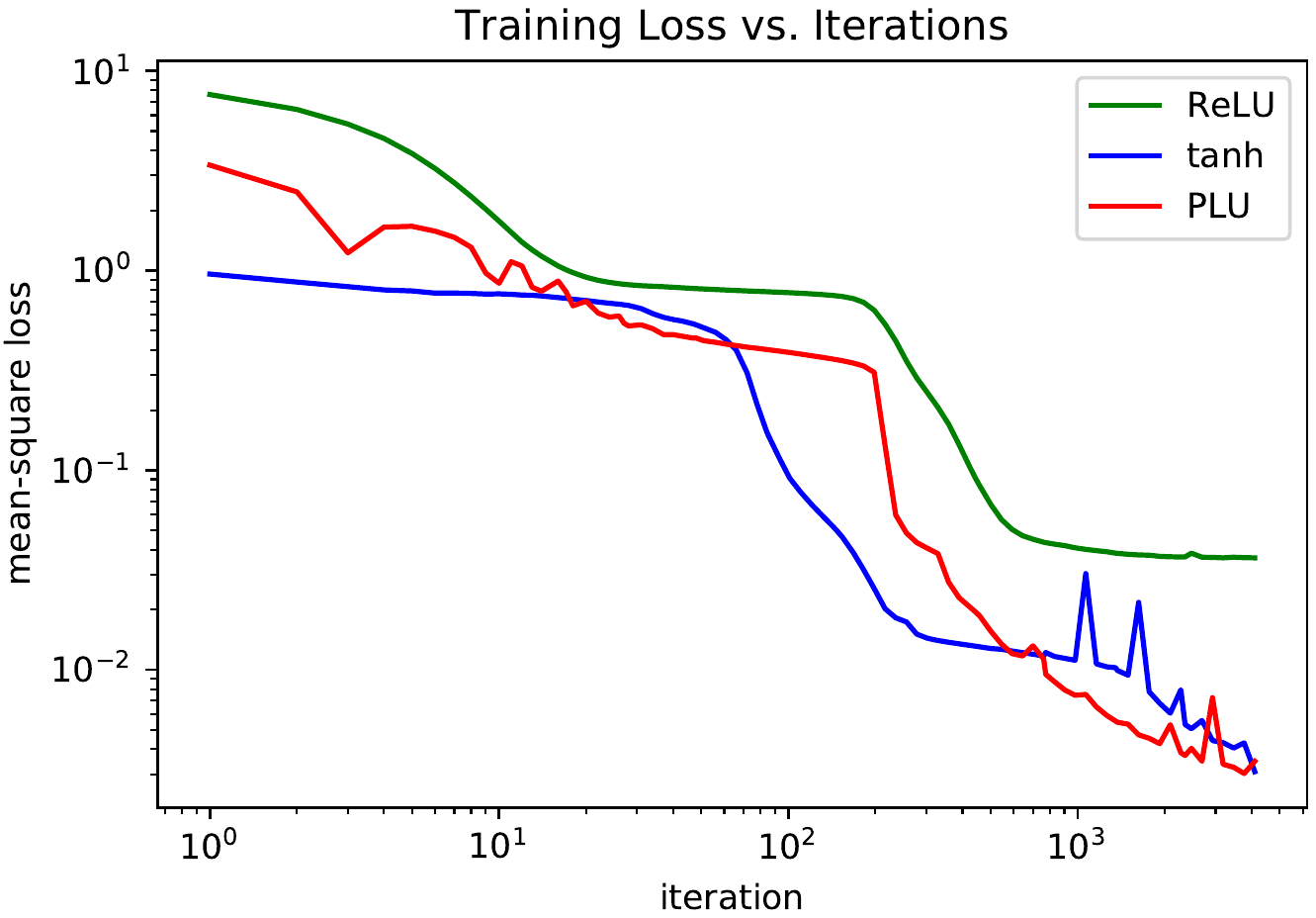}
\caption{Loss function for the 5 layer network approximating the parametric function $f(t)$.}
\label{parametricLoss}
\end{figure}

\section{Fitting 3D Surfaces}
To visualize the planar approximation to smooth surfaces in $\mathbb{R}^3$, the hyperbolic paraboloid 
\begin{equation*}
f(x,y) = x^2 - y^2
\end{equation*}
was chosen as the target function for a network model $F: \mathbb{R}^2 \to \mathbb{R}$ of depth and width 3, akin to that in section \ref{sec:sine}. The network was trained using batches of 100 uniformly drawn points $x, y \sim \mathcal{U}(-3, 3)$ with the same optimizer hyperparameters as before. 

\begin{figure}[t]
\centering
\includegraphics[width=.5\textwidth]{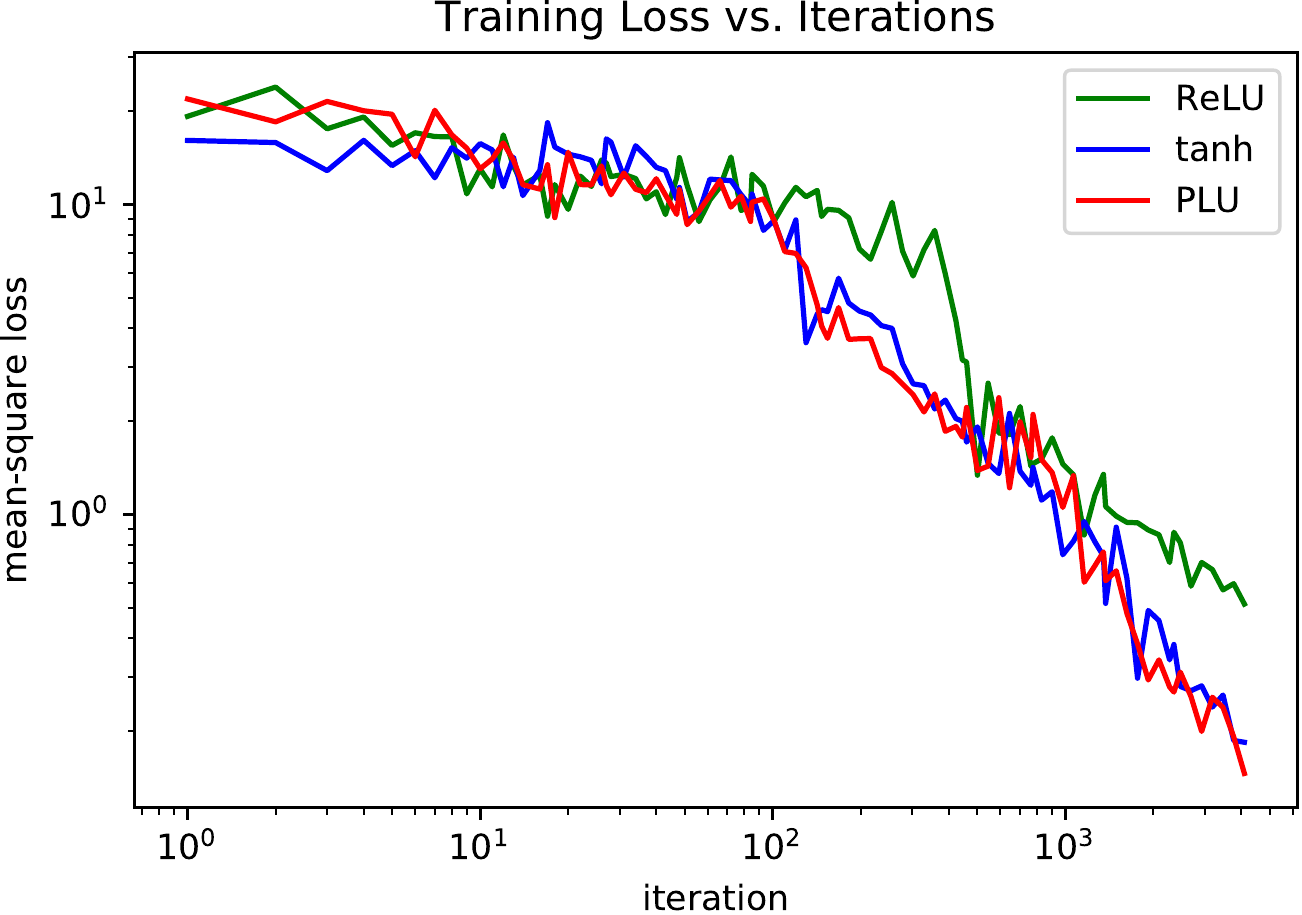}
\caption{Training loss when approximating a hyperbolic paraboloid.}
\label{hyperLoss}
\end{figure}

\begin{figure}[h!]
\centering
\includegraphics[width=.5\textwidth]{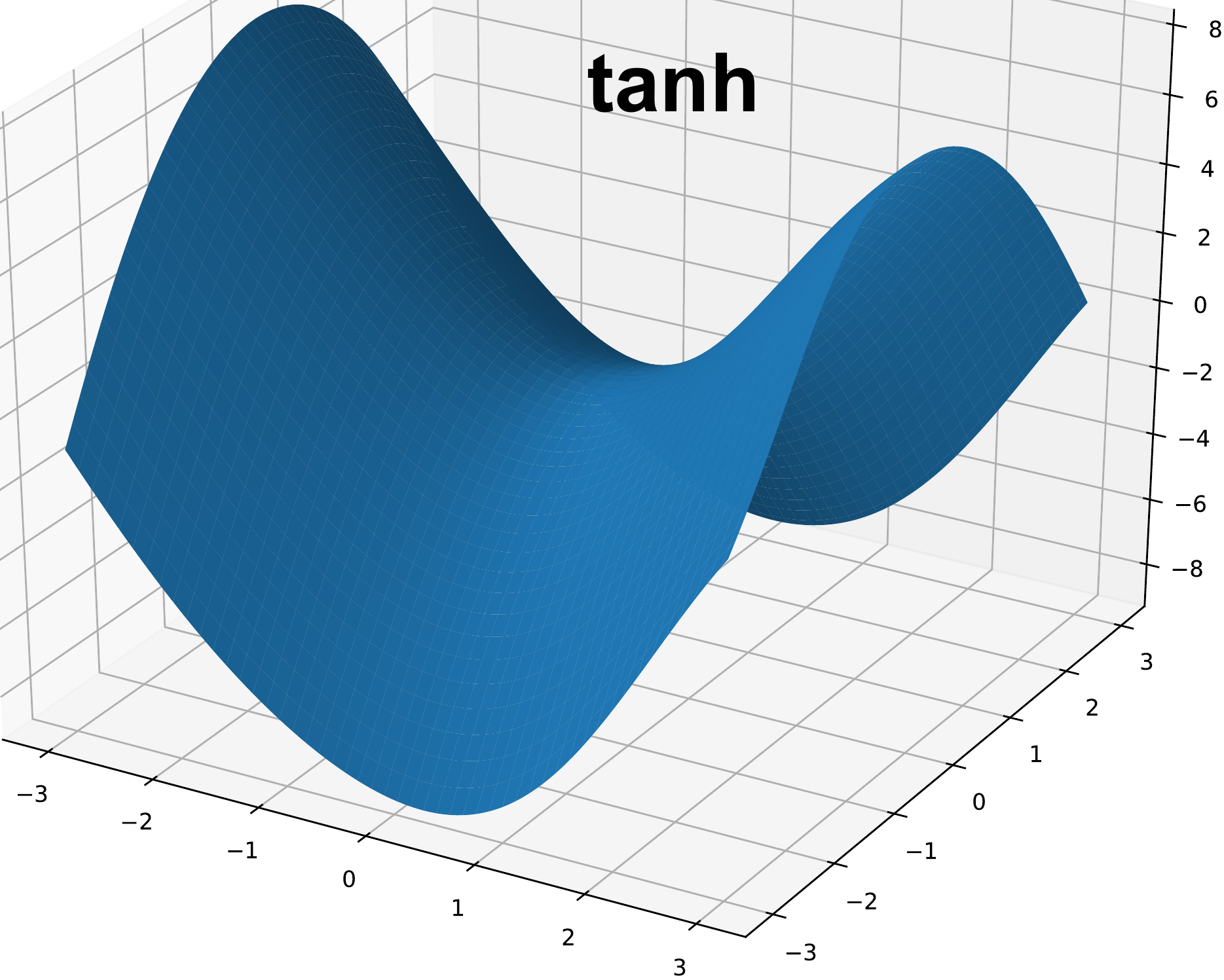}
\includegraphics[width=.5\textwidth]{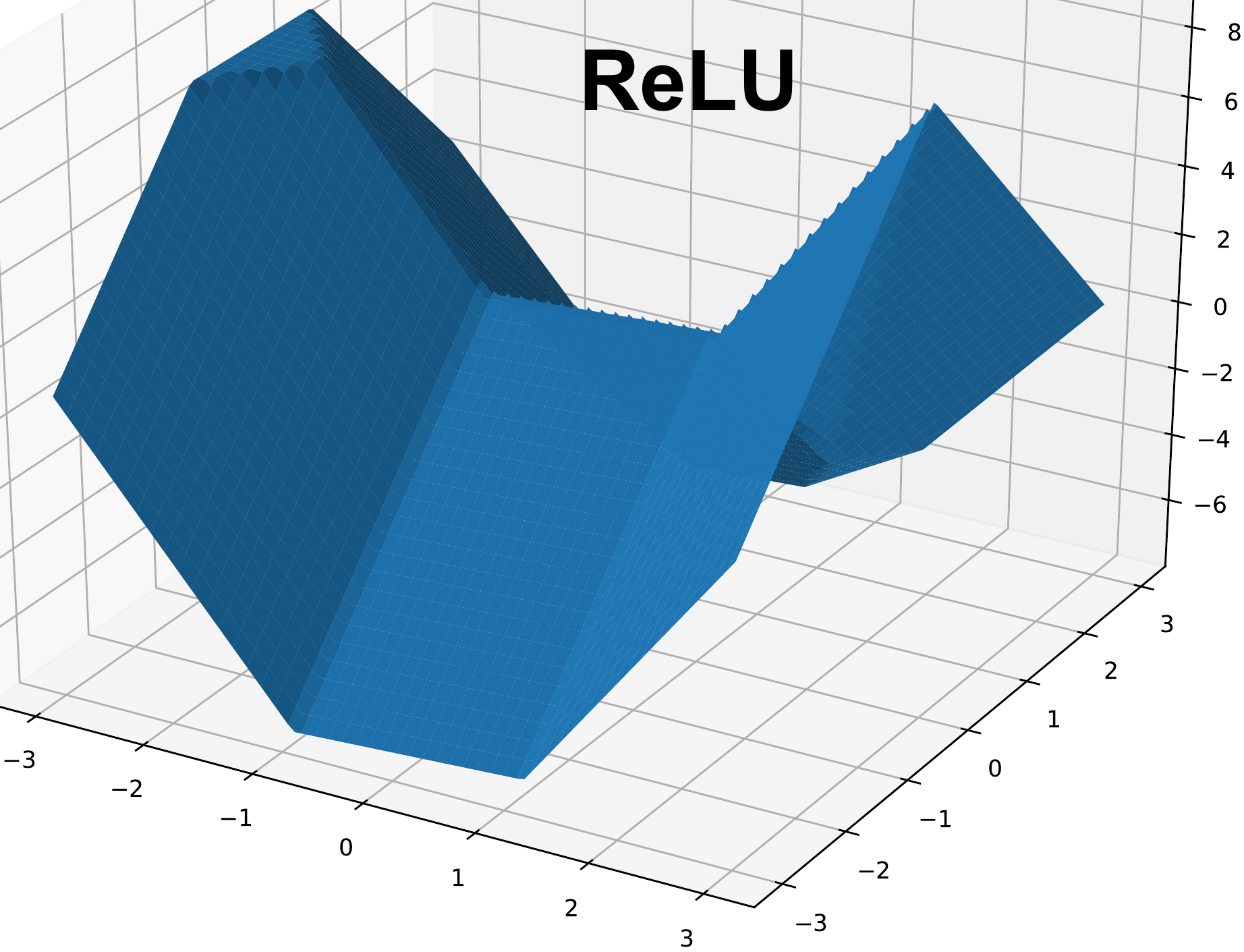}
\includegraphics[width=.5\textwidth]{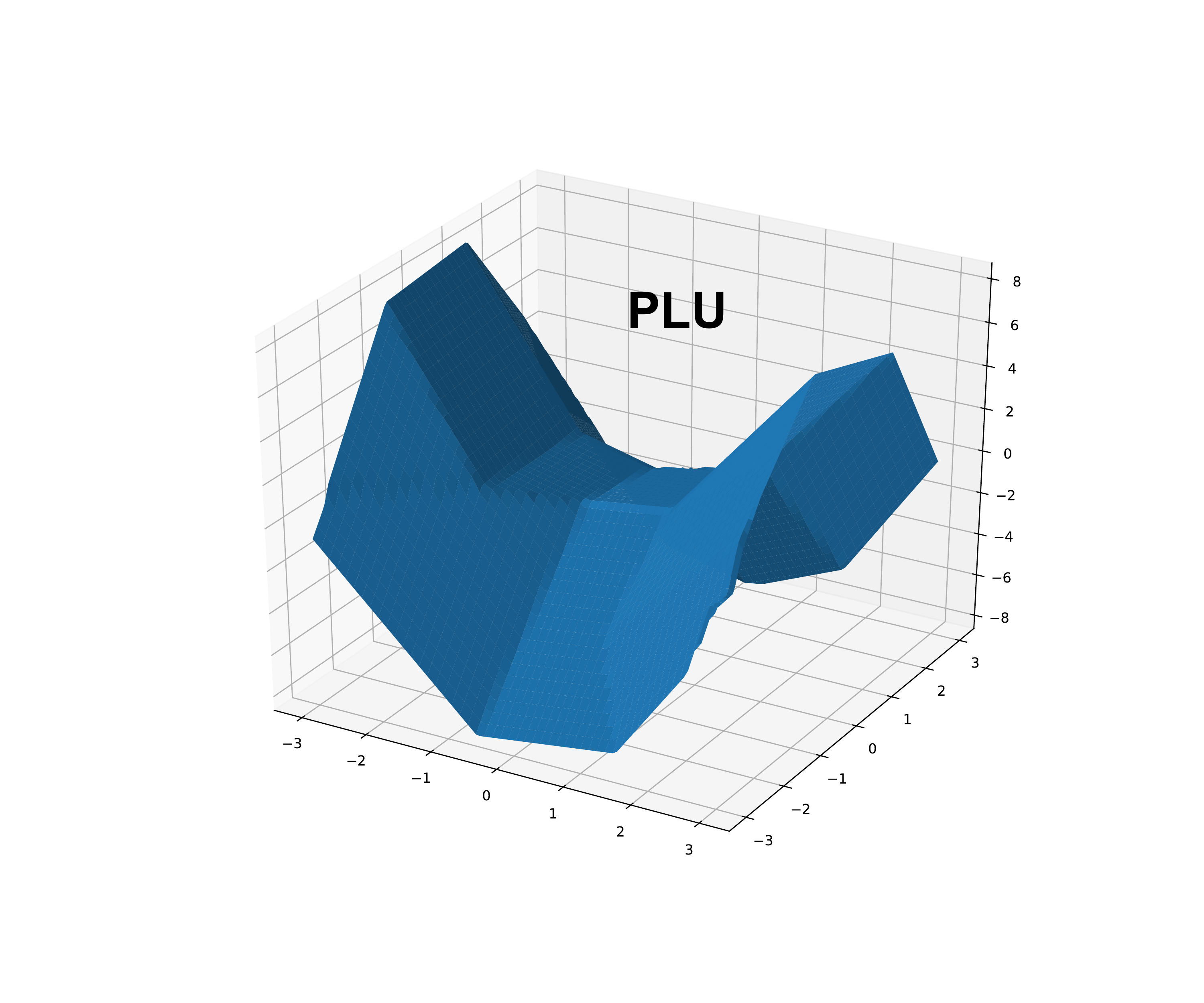}
\caption{Hyperbolic paraboloid approximations.}
\label{hyperPlots}
\end{figure}

The results parallel the story in the previous sections: the PLU outperformed the ReLU by having more available linear pieces to approximate the hyperbolic paraboloid, while the the tanh yielded a smooth surface. Figure (\ref{hyperPlots}) is an aesthetic visual of the advantage of the PLU over the ReLU and of the difference between piecewise linear and smooth activation functions. The training loss shown in figure (\ref{hyperLoss}) again shows the PLU keeping up with the tanh while the ReLU converging to a much higher loss, as evident in the function plots.

\section{CIFAR-10} \label{sec:cifar}
\begin{figure*}[t]
\center
\includegraphics[width=1.0\textwidth]{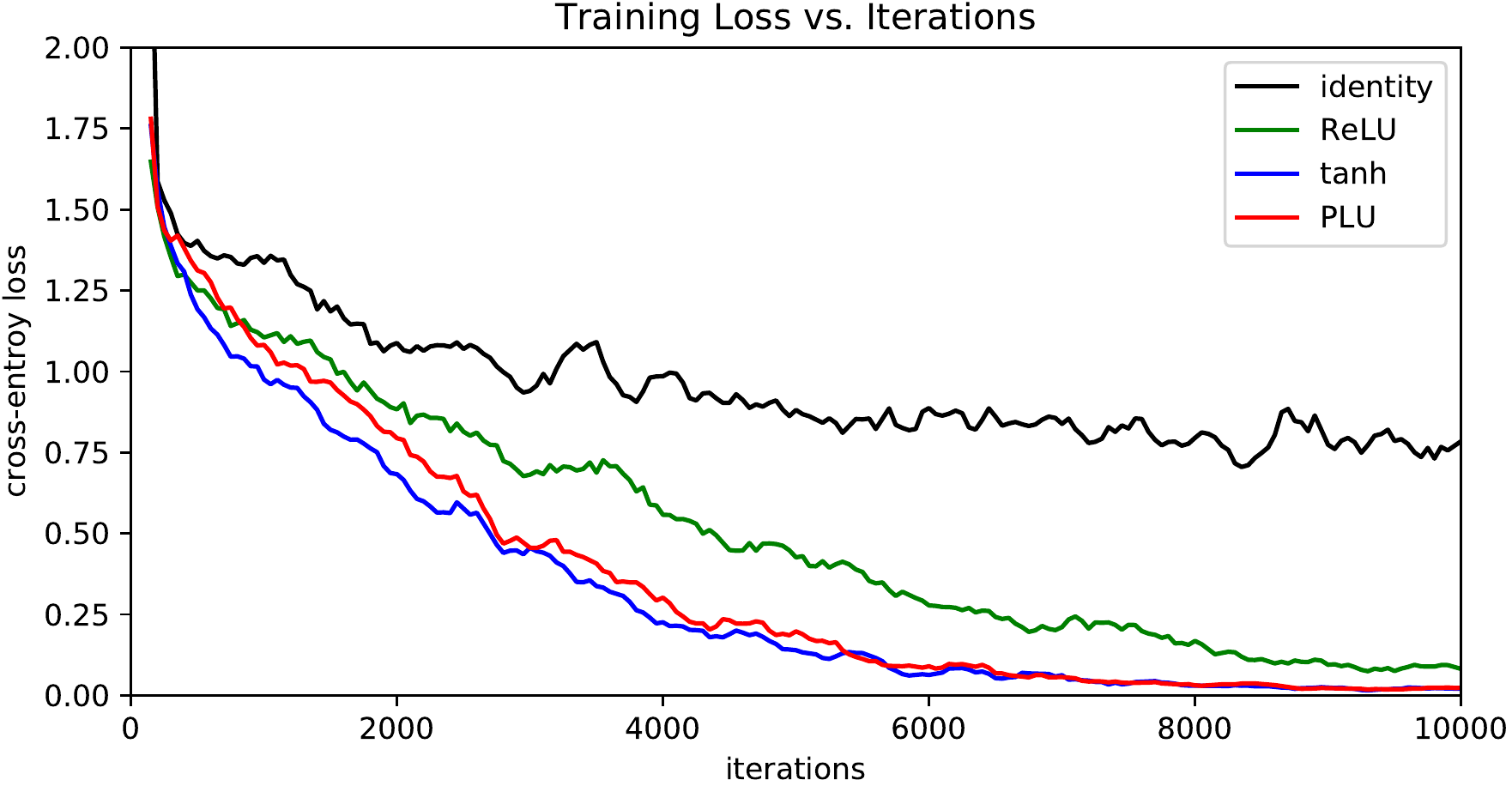}
\caption{Loss function on the CIFAR-10 dataset for 4 activation functions.}
\label{cifarLoss}
\end{figure*}
The PLU was tested with a shallow convolutional network trained on the CIFAR-10 dataset. The network architecture (table \ref{cnnTable}) was trained with TensorFlow to minimize cross-entropy loss by an Adam optimizer with 0.001 learning rate and default hyperparameters otherwise. Data was fed in batches by randomly selecting 100 images from the training set and the network parameters were optimized for $10^5$ steps. The identity function $a(x) = x$ was also used in these experiments to have a linear classifier as a datum.

\begin{table}
\centering
\begin{tabular}{l  c  c}
\textbf{Operation} & \textbf{Filters} & \textbf{Size / Stride} \\
\hline
Convolution & 16 & 3x3, s1\\
MaxPool & - & 2x2, s2 \\
Activation & - & - \\
Convolution & 32 & 3x3, s1 \\
MaxPool & - & 2x2, s2 \\
Activation & - & - \\
Convolution & 64 & 3x3, s1 \\
MaxPool & - & 2x2, s2 \\
Activation & - & - \\
Dense & - & 1024x256 \\
Activation & - & - \\
Dense & - & 256x64 \\
Activation & - & - \\
Dense & - & 64x10 \\
Softmax & - & - \\
\\
\end{tabular}
\caption{ConvNet architecture. The activation function was one of four: tanh, ReLU, PLU, identity.}
\label{cnnTable}
\end{table}

The results for this network and dataset are more favorable to the ReLU than the previous examples. One possibility is that the optimal function that maps the data to the 10 image classes of this dataset is not far from linear. The linear classifier (network with identity activation) performs well especially in the initial steps which suggests that a few nonlinearities are sufficient to adequately separate the data points. However, the tanh and PLU still train faster than the ReLU, an unsurprising result in view of the previous experiments. 

\section{Invertible Networks}
A one-to-one network function $F: \mathbb{R}^n \to \mathbb{R}^n$ can be inverted only if the activation functions are also invertible. The ReLU is not an invertible function (although the leaky-ReLU is). The tanh inverse is defined only for the open set $(-1, 1)$. The PLU is invertible with its inverse given by

\begin{equation*}
PLU^{-1} = min((x+c)/\alpha - c, max((x-c)/\alpha+c)) 
\end{equation*}
which is well defined for all $x$. Thus, the PLU offers this additional advantage of serving as an appropriate activation for constructing invertible network models.

\section{Remarks}
The piecewise linear unit, PLU, a new activation function for network models was proposed and tested against the common functions tanh and ReLU. Fitting a nonlinear function using a network model $F: \mathbb{R}^n \to \mathbb{R}^m$ relies on the nonlinearity of the activation function. The ReLU is a piecewise linear function made of two line segments, one with zero derivative. For this reason it is a poor choice of activation for fitting certain smooth nonlinear functions, as shown in this paper. The tanh is smooth and highly nonlinear, however it suffers from the vanishing gradients issue. This motivated the PLU which is a hybrid between the tanh and ReLU - a piecewise linear odd function constructed of three linear pieces that roughly approximates the tanh for a given range. The PLU was shown in several cases to outperform the ReLU significantly, mirroring the performance of the tanh rather closely albeit without the vanishing gradients issue. It also avoids the dead parameter problem of the ReLU (addressed by the leaky-ReLU). The PLU is inexpensive to compute, has nonzero derivative everywhere, and its inverse is well defined on the entire domain.

The PLU's advantage over the ReLU is most apparent when using a network model to fit highly nonlinear functions. Else, the difference between the two activation functions is diminished as suggested by the convolutional network tested in section \ref{sec:cifar}. Convolutional networks are largely feature extractors with a simple classifier in a very high dimensional space - hence the reason why HOG + SVM was so effective an algorithm. Nevertheless, the PLU is a must-try as an alternative to the ReLU especially for shallow network models and/or those attempting to fit highly nonlinear functions.

\newpage

\nocite{*}
\bibliographystyle{ieeetr}
\bibliography{PLUrefs}

\end{document}